\documentclass{article}
\usepackage{graphicx} % Required for inserting images
\usepackage[authoryear]{natbib}

\usepackage{amsmath}
\usepackage{amssymb}
\usepackage{amsthm}

\newtheorem{theorem}{Theorem}[section]
\newtheorem{proposition}[theorem]{Proposition}

\theoremstyle{definition}
\newtheorem{definition}[theorem]{Definition}

\title{Multi-Agent Constraint Factorization Reveals Latent Invariant Solution Structure}
\author{
Christopher Scofield \\
\texttt{chris@sentienta.ai}
}
\date{January 2026}

\begin{document}

\maketitle

\begin{abstract}
    Multi-agent systems (MAS) composed of large language models often exhibit improved problem-solving performance despite operating on identical information. In this work, we provide a formal explanation for this phenomenon grounded in operator theory and constrained optimization. We model each agent as enforcing a distinct family of validity constraints on a shared solution state, and show that a MAS implements a factorized composition of constraint-enforcement operators. Under mild conditions, these dynamics converge to invariant solution sets defined by the intersection of agent constraint sets. Such invariant structures are generally not dynamically accessible to a single agent applying all constraints simultaneously, even when expressive capacity and information are identical. We extend this result from exact constraint enforcement to soft constraints via proximal operators, and apply the formalism to contemporary text-based dialog systems.
\end{abstract}

\section{Introduction}

\subsection{Motivation}

Multi-agent systems (MAS) composed of large language models are increasingly used in practice. Common configurations include structured debate role- or persona-based agents, critique-revision loops, and orchestrated workflows in which multiple agents interact through a shared dialog \citep{Chen2024LLMMA}. Across a wide range of tasks, including analysis, planning, writing, and decision support, systems that generate, compare, or coordinate multiple reasoning trajectories or agent perspectives often exhibit improved performance relative to a single model queried once or even iteratively \citep{Ye2025XMASSystems}.

Despite this empirical success, existing explanations for why multi-agent systems (MAS) are effective remain incomplete. Earlier work on human collective reasoning provides an important conceptual underpinning, showing that groups can outperform individuals when problem solving benefits from diversity of perspectives, complementary heuristics, or the aggregation of partially independent judgments \citep{Surowiecki2004Wisdom,Page2007Diversity,Woolley2010Collective}. Building on these intuitions, contemporary research on MAS in machine learning has begun to explore a variety of architectures and interaction mechanisms in LLM-based multi-agent systems \citep{Tran2025MultiAgentSurvey}.

Recent work has also begun to move beyond architectural taxonomies and empirical benchmarks toward explicit analyses of interaction dynamics, stability, and population-level effects in multi-agent LLM systems.
\citep{Flint2025GroupSize}. 

However, there remains no unified mechanistic account of how agent interaction acting on a shared state alters underlying problem-solving dynamics, nor a general theory explaining why comparable benefits cannot in all cases be obtained by a single, unconstrained model with similar expressive capacity.

This gap motivates the present work: to provide a precise, mathematically grounded explanation for when and why multi-agent LLM systems succeed. Existing multi-agent LLM studies demonstrate empirical gains, but do not explain, at the level of operators, dynamics, or convergence, why interaction changes which solutions are reachable.

\subsection{Contributions}

This paper makes the following contributions:

\begin{enumerate}
\item \textbf{Agent formalization.} We formalize agents as constraint-enforcement operators acting on a shared solution state, with agent outputs interpreted as state-updating operations that impose validity conditions.

\item \textbf{Emergence via operator composition.} We prove that composing such operators yields invariant solution sets defined by the intersection of agent constraint sets, and that these sets are generally inaccessible to single-agent dynamics.

\item \textbf{Proximal optimization interpretation.} We extend the analysis from exact constraint enforcement to soft constraints using proximal operators, providing a framework compatible with approximate, incremental reasoning.

\item \textbf{Application to unstructured text systems.} We map the formalism to contemporary text-based multi-agent systems by treating dialog as system state and personas as generators of domain-specific penalties.

\end{enumerate}

\section{Related Work}

This section reviews three bodies of work that intersect in contemporary multi-agent large language model systems. These include multi-agent prompting and orchestration, classical constraint satisfaction via projections and proximal splitting, emergence and collective computation in distributed systems, and the specific theoretical gaps targeted by this work.

\subsection{Multi-Agent LLM Systems}

\paragraph{Debate and multi-agent deliberation.}
The modern debate paradigm for machine learning systems is canonically articulated in \emph{AI Safety via Debate} \citep{Irving2018Debate}, which frames two agents that argue opposing positions while a judge selects the winner. More recent work operationalizes debate as a practical inference-time improvement technique for reasoning and factuality, in which multiple LLM instances propose answers and critique one another over multiple rounds, often improving aggregate performance \citep{Du2023MultiAgentDebate}. Related structured debate methods include Multi-Agent Debate, which encourages divergent reasoning, with a judge managing the debate iterations \citep{Liang2024MAD}.

\paragraph{Role prompting, role-play, and persona-conditioned agents.}
A large fraction of deployed multi-agent systems relies on role conditioning to induce agent heterogeneity. CAMEL proposes a role-playing framework using prompting to induce autonomous cooperation between agents and to study emergent behaviors in a society of LLMs \citep{Li2023CAMEL}. Work explicitly analyzing role-play prompting for improved Chain-of-Thought (CoT) reasoning includes \emph{Better Zero-Shot Reasoning with Role-Play Prompting} \citep{Kong2023RolePlay}.

\paragraph{Self-consistency and ensemble-style inference.}
While not always framed as multi-agent, self-consistency is a closely related inference mechanism in which multiple chains of thought are sampled and aggregated to select a consistent answer \citep{Wang2023SelfConsistency}. This approach can be interpreted as exploring multiple trajectories in a latent reasoning space followed by projection under an implicit aggregation rule.  Tree-of-Thoughts (ToT) generalizes chain-of-thought prompting into an explicit search over thought units.  ToT LLMs also explore multiple reasoning trajectories but with enhanced analysis through look-ahead and backtracking \citep{Yao2023TreeOfThoughts}. 

\paragraph{Agentic tool use and orchestration frameworks.}
Several lines of work position LLMs as agents that act through tool use in addition to producing text. ReAct interleaves reasoning traces with actions to reduce hallucination and improve task completion \citep{Yao2022ReAct}. Reflexion introduces verbal feedback and episodic text memory to improve subsequent trials without parameter updates \citep{Shinn2023Reflexion}.

At the systems level, orchestration frameworks formalize multi-agent conversation patterns and role assignment. AutoGen emphasizes programmable conversation patterns among LLM agents and tools \citep{Wu2023AutoGen}. Domain-specific collaboration frameworks include ChatDev and MetaGPT, which encode structured multi-role workflows and reduce cascading errors via explicit intermediate artifacts and procedure-like prompting \citep{Qian2023ChatDev,Hong2023MetaGPT}.

\paragraph{Learned coordination and cognitive alignment.}
Recent work has explored learned mechanisms for coordinating multi-agent LLM collaboration through adaptive communication and internal state modeling. Zhang et al.\ propose OSC, a framework in which agents maintain models of their collaborators' cognitive states and use learned communication policies to reduce representational gaps during interaction \citep{Zhang2025OSC}. This approach emphasizes adaptive alignment and efficiency in multi-agent collaboration and demonstrates empirical performance gains across cooperative tasks. OSC treats interaction as a finite-round learned communication protocol but does not provide a theoretical analysis of convergence, invariant sets, or shared-state dynamics under repeated interaction.

\paragraph{Positioning relative to this paper.}
The above work demonstrates that multi-agent systems often improve performance empirically \citep{
Wang2023SelfConsistency,
Yao2023TreeOfThoughts,
Wei2022ChainOfThought,
Du2023MultiAgentDebate,
Li2023CAMEL,
Qian2023ChatDev,
Huang2025AdaptiveCoopetition,
Ye2025XMASSystems}. The contribution of the present work is to supply a mathematically explicit account of when and why role-conditioned, interacting LLM instances can access invariant solution subspaces that are not reliably reached by a single unconstrained inference trajectory.

\subsection{Constraint Satisfaction and Alternating Projections}

\paragraph{Von Neumann alternating projections.}
A classical prototype for iterative constraint enforcement is alternating orthogonal projection between two closed subspaces of a Hilbert space. Von Neumann showed that, when the intersection is nonempty, repeated application of the projection operators converges to the projection of the initial point onto the intersection \citep{VonNeumann1933Projections}. This result illustrates how a joint solution can be recovered through the sequential enforcement of partial constraints rather than by direct joint optimization. Subsequent work has extended this framework beyond linear subspaces to more general constraint sets, clarifying both convergence behavior and failure modes when projections are approximate or constraints are incompatible \citep{Hundal2004Alternating,Lewis2008AlternatingManifolds}.

\paragraph{Douglas-Rachford and monotone operator splitting.}
Douglas-Rachford splitting originates in numerical methods for partial differential equations \citep{Douglas1956Rachford} and was later understood as a general splitting method for finding zeros of sums of monotone operators. Lions and Mercier developed splitting algorithms for sums of nonlinear monotone operators, providing an operator-theoretic foundation \citep{Lions1979Splitting}.

\paragraph{ADMM and augmented Lagrangian methods.}
The alternating direction method of multipliers (ADMM) arises from augmented Lagrangian decomposition methods developed in the 1970s for constrained convex optimization. A widely used synthesis by Boyd et al.\ reframes ADMM explicitly as a decomposition and coordination algorithm, emphasizing its suitability for large-scale and distributed optimization problems in which variables and constraints can be partitioned across subsystems \citep{Boyd2011ADMM}. ADMM provides an important precedent for understanding how sequential enforcement of partial constraints can yield globally coherent solutions without centralized optimization.

\paragraph{Proximal point and proximal splitting.}
Rockafellar formalized the proximal point algorithm for maximal monotone operators \citep{Rockafellar1976Proximal}. Eckstein and Bertsekas established relationships between Douglas-Rachford splitting, proximal point methods, and ADMM \citep{Eckstein1992DR}. Comprehensive treatments of proximal operators and splitting algorithms are provided by Parikh and Boyd and by Combettes and Pesquet \citep{Parikh2014Proximal,Combettes2011Splitting}.

\paragraph{Positioning relative to this paper.}
The present work leverages these classical results but introduces a novel interpretation. Persona-conditioned LLM responses are modeled as approximate projection or proximal updates in a text-induced state space. Multi-agent interaction corresponds to composition of these operators, with convergence to invariant sets defined by joint constraint satisfaction rather than optimization of a single monolithic objective.

\subsection{Emergence and Collective Computation}

\paragraph{Distributed optimization and consensus.}
In distributed optimization, agents maintain local objective functions or data, and coordination arises through iterative communication and local updates. Classical distributed subgradient and consensus-based methods provide convergence guarantees under minimal information sharing, typically without any agent acquiring new external data beyond messages exchanged with neighbors \citep{Nedic2009Distributed}. Asynchronous variants of such distributed computations have also been studied extensively, demonstrating robustness to delays and partial updates \citep{OlfatiSaber2007Consensus, Tsitsiklis1986Distributed}. In contrast to the operator-theoretic approach developed here, the framework of \citet{Nedic2009Distributed} assumes a fixed global optimization objective decomposed across agents, rather than agents whose interactions dynamically transform internal representations and intermediate operators during execution.

\paragraph{Population-level dynamics and group size effects.}
Flint et al.\ analyze group size effects in LLM-based multi-agent systems using a dynamical-systems framework, deriving a mean-field description in which, above a critical population size, stochastic interactions concentrate around deterministic trajectories with stable fixed points \citep{Flint2025GroupSize}. Their results establish group size as a control parameter governing transitions between fluctuation-dominated and stable collective regimes, providing a clear population-level account of emergent coordination. While their analysis focuses on consensus and absorbing states of a population process, our work addresses a complementary question: how interaction between heterogeneous, role-conditioned agents alters invariant-set accessibility through composition of agent-specific operators acting on a shared problem state.

\paragraph{Collective intelligence and diversity-based advantage.}
Empirical work on collective intelligence studies whether group performance exceeds what is predicted by individual capability alone. Woolley et al.\ report evidence for a collective intelligence factor in human groups \citep{Woolley2010Collective}. Closely related but analytically distinct, Hong and Page present a formal agent-based model in which heterogeneous problem solvers, endowed with different heuristics and representations, collectively outperform groups composed solely of individually high-performing agents \citep{Hong2004Diversity}. Their result establishes, under bounded rationality assumptions, that diversity alone can yield systematic performance advantages even in the absence of learning or information sharing.

\subsection{Gaps in Existing Theory}

Existing multi-agent LLM systems demonstrate empirical gains through patterns such as debate or role specialization, but typically do not formalize the operators induced by agent updates or characterize the invariant solution structures reachable under repeated interaction. Classical optimization and consensus frameworks provide such characterizations for explicitly defined operators on structured state spaces, but do not directly apply to stochastic, text-mediated updates. Conversely, operator-theoretic approaches rigorously analyze convergence and invariant sets under operator composition, but are developed for algebraically or geometrically structured settings and do not model persona-conditioned natural-language interaction. This paper bridges these perspectives by introducing an operator-level formalism in which agent responses act as approximate constraint-enforcement operators on a shared state encoded in text, enabling analysis of invariant-set accessibility without assuming a single global objective or centralized optimization.

\section{Problem Setup and Notation}

\subsection{Solution Space and State}

We consider a fixed problem specified by an initial prompt or query. Let \(\mathcal{S}\)
denote the space of all candidate solutions consistent with this specification. The space \(\mathcal{S}\) is defined abstractly and may be large or implicitly specified; we do not assume that its elements are enumerated or explicitly represented. In the systems of interest, elements of \(\mathcal{S}\) correspond to semantic dialog states, such as evolving sequences of natural language exchanges.

For purposes of analysis, we associate each dialog state with a \emph{state}
\[
x \in \mathcal{X},
\]
where \(\mathcal{X}\) is a real Hilbert space equipped with inner product \(\langle \cdot, \cdot \rangle\) and induced norm \(\lVert \cdot \rVert\). The state \(x\) represents a partial or provisional representation of a solution to the query, obtained by encoding the current dialog state. In later sections, \(\mathcal{X}\) will be instantiated both as a finite-dimensional vector space and as a structured representation derived from unstructured text.

\subsection{Agents and Personas}

We consider a collection of \(m\) agents indexed by \(i \in \{1, \dots, m\}.\)
Each agent is associated with a \emph{persona}, which specifies a domain of expertise or evaluative focus. Importantly, personas do not restrict the expressive capacity of an agent. Instead, they determine which validity constraints the agent is authorized to apply when evaluating or modifying the shared state.

We assume that all agents have access to the same information: the initial problem specification (the query) and the evolving shared state. No agent possesses private data or privileged observations.

\subsection{Constraint Sets and Feasible Regions}

Each agent \(i\) induces a set of validity constraints on the solution space. These constraints define a subset
\[
A_i \subset \mathcal{X},
\]
interpreted as the set of states that satisfy all constraints expressible within agent \(i\)'s domain of expertise.

The sets \(A_i\) are not assumed to be explicitly enumerated. Rather, they are defined implicitly through the agent's evaluative behavior. In the idealized setting, we assume the following:
\begin{itemize}
\item each \(A_i\) is closed and convex,
\item the collective feasible set
\[
A \equiv \bigcap_{i=1}^m A_i
\]
is non-empty.
\end{itemize}

The set \(A\) represents states that simultaneously satisfy all agent-imposed constraints and serves as the candidate invariant solution set for the multi-agent dynamics. In this sense, each \(A_i\) and their intersection \(A\) may be viewed as feasible regions within the representation space \(\mathcal{X}\).

\subsection{Agent Update Operators}

Each agent \(i\) acts on the shared state by applying an update operator
\[
T_i : \mathcal{X} \to \mathcal{X},
\]
which maps the current state to a revised state reflecting the agent's evaluation and modification of the solution.

In the idealized setting, \(T_i\) is assumed to act as a constraint-enforcement operator associated with the feasible set \(A_i\). In particular, \(T_i(x)\) moves the state toward consistency with the constraints defining \(A_i\), without introducing new external information. We do not assume that \(T_i\) is an exact projection onto \(A_i\); rather, it may be stochastic, approximate, or history-dependent, reflecting the behavior of large language models operating on unstructured text.

When convenient, \(T_i\) may be interpreted as an approximate projection or proximal update with respect to \(A_i\), but the analysis that follows depends only on its role as an operator whose fixed points lie in \(A_i\).

\section{Constraint Enforcement and Operator Factorization}

We now analyze the composition of agent update operators and its implications for constraint satisfaction and invariant-set structure.

\subsection{Agent Operators as Constraint Enforcement}

For the analysis that follows, we assume that operators \(T_i\) satisfy the following properties.

\paragraph{Fixed points on feasible sets.}
States that satisfy agent \(i\)'s constraints are fixed points of \(T_i\). Accordingly, the feasible set \(A_i\) coincides with the fixed-point set of the operator.

\paragraph{Constraint-correcting behavior.}
For states outside the feasible set, \(T_i\) acts to reduce violation of agent \(i\)'s constraints, modeling the agent's proposal of revisions that move the state toward admissibility.

\paragraph{Non-expansiveness.}
Each operator \(T_i\) is assumed to be non-expansive. This regularity condition ensures stability of the induced dynamics and rules out pathological updates that would destabilize iterative constraint enforcement.

\subsection{Idealized Case: Projection Operators}

In the idealized setting, each agent operator is taken to be the orthogonal projection onto its feasible set:
\[
T_i(x) = P_{A_i}(x),
\]
where
\[
P_{A_i}(x) = \operatorname*{arg\,min}_{z \in A_i} \lVert z - x \rVert,
\]

Under this model, each agent minimally modifies the shared state so that it satisfies the agent's constraints, without introducing additional structure or preferences. Although exact projections are not intended as literal models of LLM behavior, they provide a mathematically transparent baseline for analyzing the effects of sequential constraint enforcement.

\subsection{Sequential Interaction and Shared State}

Agents act sequentially on a single shared state. One interaction round applies the composed operator
\[
T \equiv T_m \circ T_{m-1} \circ \cdots \circ T_1.
\]
It is this composition, rather than the individual operators \(T_i\), that governs the evolution of the system across interaction rounds.

\subsection{Why Operator Factorization Matters}

The distinction between a single-agent and a multi-agent system can now be stated precisely.

\paragraph{Single-agent dynamics.}
A single-agent system applies a single operator that attempts to enforce all constraints simultaneously. A multi-agent system applies a factorized sequence of operators, each enforcing a subset of constraints.

\paragraph{Multi-agent dynamics.}
As we will show in the next section, this difference is not merely procedural. Under mild conditions, the composition of constraint-enforcement operators reveals invariant solution structures, corresponding to intersections of feasible sets, that are not dynamically accessible to any single operator acting alone.  This observation forms the core theoretical explanation for the effectiveness of multi-agent systems.

\section{Emergence via Operator Composition}

In this section we give a precise operator-theoretic account of emergence in multi-agent systems. We show that factorizing constraint enforcement across agents induces invariant solution structures that are not dynamically accessible to individual agents or to monolithic, non-factored updates, even when all agents have identical information.

\subsection{Invariant Sets Induced by Factored Dynamics}

Let \(\mathcal{X}\) be a real Hilbert space, and let \(\{A_i\}_{i=1}^m\) be nonempty, closed, convex subsets of \(\mathcal{X}\) with nonempty intersection
\[
A \equiv \bigcap_{i=1}^m A_i.
\]
Let each agent operator be the orthogonal projection \(T_i = P_{A_i}\), and define the composed multi-agent operator
\[
T \equiv P_{A_m} \circ P_{A_{m-1}} \circ \cdots \circ P_{A_1}.
\]

\begin{theorem}[Emergent Invariant Set via Factored Projections]
For any initial state \(x^{(0)} \in \mathcal{X}\), the iterates
\[
x^{(k+1)} = T\bigl(x^{(k)}\bigr)
\]
are Fej\'er monotone with respect to \(A\), i.e.,
\[
\lVert x^{(k+1)} - a \rVert \le \lVert x^{(k)} - a \rVert \quad \forall a \in A,\ \forall k \ge 0.
\]
Consequently, \(A\) is an invariant set of the multi-agent dynamics, and every weak cluster point of \(\{x^{(k)}\}\) lies in \(A\).
\end{theorem}

\paragraph{Proof.}
Each orthogonal projection \(P_{A_i}\) is firmly non-expansive. Standard results on cyclic projections onto closed convex sets imply that the composed operator \(T\) is non-expansive and that the iterates are Fej\'er monotone with respect to the intersection \(A\) \citep{BauschkeCombettes2017}. Fej\'er monotonicity implies boundedness, monotone decrease of the distance to \(A\), and that every weak accumulation point lies in \(A\). \(\square\)

\paragraph{Interpretation.}
The multi-agent system dynamically stabilizes on the intersection of agent constraint sets. Importantly, no agent enforces the full constraint family defining \(A\); the invariant structure arises only through sequential operator composition.

\subsection{Non-Representability by Individual Agents}

We now formalize the sense in which the emergent invariant structure cannot be accessed by any individual agent acting alone.

\begin{proposition}[Strict Emergence]
If \(A \subsetneq A_i\) for all \(i \in \{1,\dots,m\}\), then \(A\) is not the fixed-point set of any individual operator \(P_{A_i}\).
\end{proposition}

\paragraph{Proof.}
The fixed-point set of \(P_{A_i}\) is exactly \(A_i\). Since \(A \subsetneq A_i\), no individual operator admits \(A\) as its invariant set. \(\square\)

Thus, even though each agent acts deterministically and has full access to the shared state, no single agent can dynamically enforce the collective solution structure.

\subsection{Degenerate Cases: Absence of Emergence}

Emergence disappears when constraint factorization collapses.

\begin{proposition}[Identical Agents]
If \(A_1 = A_2 = \cdots = A_m\), then
\[
T = P_{A_1},
\]
and the multi-agent system reduces to a single-agent projection.
\end{proposition}

In this degenerate case, interaction introduces no new invariant structure. This corresponds to systems in which all agents share identical personas, evaluative authority, and constraint sets.

\subsection{Why a Single Monolithic Agent Can Fail}

One might object that a sufficiently capable single agent could enforce all constraints simultaneously by applying the joint projection \(P_A\) onto
\[
A = \bigcap_{i=1}^m A_i.
\]
If such an operator were available, factorization would indeed offer no advantage.

In practice, however, a single agent enforces multiple constraints through a coupled trade-off rather than exact projection. Regardless of the complexity of its internal reasoning, its externally observable behavior induces a single update operator
\[
S : \mathcal{X} \to \mathcal{X},
\]
corresponding to the resolution of competing constraint penalties. A convenient abstraction, formalized in the next section, is a regularized update of the form
\[
S(x) = \operatorname{prox}_{\sum_{i=1}^m \lambda_i \phi_i}(x),
\]
where each \(\phi_i\) penalizes violation of constraints associated with \(A_i\).

When constraints are enforced via penalties rather than projections, violations may be tolerated if they reduce other penalties. Constraint satisfaction becomes negotiable, and invariance of the full intersection \(A\) is no longer guaranteed. Accordingly, the fixed points of \(S\) correspond to trade-off optima rather than hard-feasible solutions. These optima may lie outside \(A\), or may collapse a high-dimensional feasible region into a single preferred point.

By contrast, the factored operator
\[
T = P_{A_m} \circ \cdots \circ P_{A_1}
\]
admits \(A\) itself as an invariant set and preserves its geometric structure under iteration. The advantage of a multi-agent system is therefore not increased expressive power, but the structural enforcement of constraints through operator factorization rather than monolithic coupling.

\subsection{Summary}

This section shows that factorizing constraint enforcement across agents fundamentally alters the invariant structure of the dynamics. While individual agents cannot stabilize the collective feasible set \(A\), compositions of projection operators admit \(A\) itself as an invariant set and preserve its geometric structure under iteration. By contrast, monolithic updates that couple constraints through trade-offs need not preserve feasibility or invariance, even when operating over the same information. Emergence in this setting arises from changes in dynamical accessibility, not from an expansion of the solution space.

\section{Robustness to Approximate Constraint Enforcement}

The analysis in Section~5 assumes exact constraint enforcement via orthogonal projections. While this idealization provides theoretical clarity, it does not reflect how large language model agents operate in practice. Agent updates are typically approximate, incremental, and soft. In this section, we show that the emergence results of Section~5 persist under such approximate enforcement.

\subsection{Soft Constraints and Penalty Functions}

Instead of representing agent constraints solely by feasible sets \(A_i\), we associate each agent \(i\) with a penalty function
\[
\phi_i : \mathcal{X} \to [0,\infty],
\]
where \(\phi_i(x)\) measures violation of agent \(i\)'s validity conditions. We assume that each \(\phi_i\) is proper, convex, and lower semicontinuous, and that
\[
\phi_i(x) = 0 \quad \text{if and only if} \quad x \in A_i.
\]

The global energy of the system is defined as
\[
F(x) \equiv \sum_{i=1}^m \phi_i(x).
\]

\subsection{Proximal Agent Updates}

Agent updates are modeled via proximal operators.

\begin{definition}[Proximal Operator]
Given \(\lambda > 0\), the proximal operator associated with \(\phi_i\) is
\[
\operatorname{prox}_{\lambda \phi_i}(x)
=
\arg\min_{z \in \mathcal{X}}
\left[
\phi_i(z) + \frac{1}{2\lambda} \lVert z - x \rVert^2
\right].
\]
\end{definition}

If \(\phi_i\) is the indicator function of \(A_i\),
\[
\phi_i(z) =
\begin{cases}
0, & z \in A_i, \\
+\infty, & z \notin A_i,
\end{cases}
\]
then the minimization is restricted to \(z \in A_i\), and
\(\operatorname{prox}_{\lambda \phi_i}(x)\) reduces to the orthogonal projection
\(P_{A_i}(x)\).

\subsection{Proximal Multi-Agent Dynamics}

A single multi-agent iteration takes the form
\[
x^{(k+1)}
=
\operatorname{prox}_{\lambda_k \phi_m}
\circ
\operatorname{prox}_{\lambda_k \phi_{m-1}}
\circ \cdots \circ
\operatorname{prox}_{\lambda_k \phi_1}
\bigl(x^{(k)}\bigr),
\]
where \(\{\lambda_k\}_{k \ge 1}\) is a sequence of positive step sizes.  The step sizes \(\lambda_k\) are assumed to decrease with \(k\), for example \(\lambda_k = 1/k\), and are chosen to satisfy standard summability conditions discussed below.

\subsection{Convergence Result}
\begin{theorem}[Convergence of Proximal Multi-Agent Dynamics]
Let
\[
F(x) := \sum_{i=1}^m \phi_i(x).
\]
Assume that:
\begin{enumerate}
\item each \(\phi_i\) is proper, convex, and lower semicontinuous,
\item the minimizer set \(\arg\min_x F(x)\) is non-empty,
\item the step sizes satisfy
\[
\sum_{k=1}^\infty \lambda_k = \infty,
\qquad
\sum_{k=1}^\infty \lambda_k^2 < \infty.
\]
\end{enumerate}
Then the cyclic proximal iteration weakly converges to a minimizer:
\[
x^{(k)} \rightharpoonup x^\star \in \arg\min_x F(x).
\]
\end{theorem}

This result is a standard consequence of convergence guarantees for cyclic proximal point methods under diminishing step sizes \citep{BauschkeCombettes2017}, and is included here to demonstrate robustness of the emergence mechanism beyond exact projection dynamics.

Thus, factored proximal dynamics stabilize an invariant solution set, whereas monolithic, coupled updates may select a single compromise point even when multiple feasible solutions exist.

\subsection{Summary}

This section shows that the emergence of invariant solution structures in multi-agent systems does not rely on exact projections. Sequential, approximate constraint enforcement via proximal operators preserves convergence and accessibility properties, demonstrating that the core results of Section~5 are robust to realistic agent behavior.

\section{Explicit Multi-Agent Example}

We now present a simple, fully explicit example that illustrates how factorized constraint enforcement yields an emergent invariant solution. The example is intentionally low-dimensional and analytically tractable, allowing the multi-agent dynamics to be understood in closed form.

\subsection{State Space and Agent Penalties}

Let the representation space be
\[
\mathcal{X} = \mathbb{R}^2,
\]
with state vector \(x = (x_1, x_2)\). We consider three agents, each enforcing a distinct quadratic penalty:
\begin{align*}
\phi_1(x) &= \tfrac{1}{2}(x_1 - 1)^2, \\
\phi_2(x) &= \tfrac{1}{2}(x_2 - 1)^2, \\
\phi_3(x) &= \tfrac{1}{2}(x_1 + x_2 - 1)^2.
\end{align*}

Each agent constrains a different linear projection of the state. No individual agent specifies the full solution.

The collective energy function is
\[
F(x)
=
\tfrac{1}{2}(x_1 - 1)^2
+
\tfrac{1}{2}(x_2 - 1)^2
+
\tfrac{1}{2}(x_1 + x_2 - 1)^2.
\]

\begin{proposition}
The function \(F\) has a unique minimizer
\[
x^\star = \left(\tfrac{2}{3}, \tfrac{2}{3}\right).
\]
This point is not the minimizer of any individual penalty \(\phi_i\).
\end{proposition}

As shown below, this point arises as the unique fixed point of the cyclic composition of the agents' proximal updates.

\subsection{Proximal Agent Updates}

Each agent applies a proximal update of the form
\[
x^{(k+1)} = \operatorname{prox}_{\lambda \phi_i}(x^{(k)}).
\]
Because each \(\phi_i\) is quadratic, the proximal operators admit closed-form expressions. For example,
\[
\operatorname{prox}_{\lambda \phi_1}(x)
=
\left(
\frac{x_1 + \lambda}{1 + \lambda},
\; x_2
\right),
\]
with analogous expressions for \(\phi_2\) and \(\phi_3\).

A single multi-agent iteration applies the three proximal operators sequentially in a fixed order.

\subsection{Convergence and Emergence}

Because each proximal update is an affine contraction, the cyclic composition of the three operators defines a single contractive map with a unique fixed point. Direct calculation shows that this fixed point coincides with the minimizer identified above,
\[
x^{(k)} \;\longrightarrow\; x^\star = \left(\tfrac{2}{3}, \tfrac{2}{3}\right).
\]

\paragraph{}This convergence occurs despite the fact that:
\begin{itemize}
\item no agent optimizes the global objective \(F\),
\item no agent has access to all constraints simultaneously,
\item no agent computes the full gradient of \(F\).
\end{itemize}

The limiting solution is therefore not produced by any individual agent, but emerges as the unique fixed point of the composed operator.

\subsection{Failure of Averaging and Monolithic Optimization}

For comparison, consider two alternative strategies.

\paragraph{Averaging}
Averaging the individual agent optima yields a point that does not satisfy all constraints simultaneously and does not coincide with \(x^\star\).

\paragraph{Single-Agent Monolithic Updates}

A single agent attempting to resolve all constraints simultaneously can be modeled as performing a monolithic regularized minimization,
\[
x \;\mapsto\; \arg\min_x \bigl(F(x) + \mu \|x\|^2\bigr),
\qquad \mu > 0.
\]
In this example, such regularization selects the compromise point
\[
x_\mu = \left(\frac{2}{3+\mu},\,\frac{2}{3+\mu}\right),
\]
which differs from the emergent solution
\[
x^\star = \left(\tfrac{2}{3},\,\tfrac{2}{3}\right)
\]
for all \(\mu > 0\).

The added regularization term introduces a bias toward minimal norm, continuously pulling the solution away from the emergent point and toward the origin as \(\mu\) increases. Thus, although the same constraint information is present, monolithic optimization converges to a different limit, while operator factorization yields the emergent solution \(x^\star\).

\section{Extension to Text-Based Multi-Agent Systems}

We now relate the operator-theoretic framework to contemporary multi-agent systems operating on unstructured text. The claim is not that large language models explicitly compute projections or proximal operators, but that their observable interaction dynamics can be modeled as an approximate realization of factorized constraint enforcement.

\subsection{Dialog as Shared State}

In text-based multi-agent systems, the shared system state is the evolving dialog. For analytical purposes, we associate each dialog state with an implicit representation
\[
x = E(\text{dialog}),
\]
where \(E\) maps the dialog transcript to a point in an abstract representation space \(\mathcal{X}\). This encoding need not be explicit or invertible, and agents do not observe \(x\) directly. Instead, agents read and modify the dialog, thereby inducing changes in the underlying state representation.

From the perspective of the theory, the dialog therefore plays the role of the shared state \(x^{(k)}\) in the multi-agent dynamics.

\subsection{Agent Turns as Factorized Updates}

Each agent is associated with a persona that determines a domain of evaluative authority - i.e., which types of constraints the agent is permitted to assess and address. Given a dialog state, an agent produces a response that attempts to reduce violations of constraints within its domain. At an abstract level, this behavior can be modeled as an approximate constraint-enforcement update
\[
x^{(k+1)} \approx T_i(x^{(k)}),
\]
where \(T_i\) is an operator associated with agent \(i\).

Operationally, this update is realized through text generation. Because language generation is discrete, context-dependent, and stochastic, these operators are noisy and inexact. Nevertheless, their sequential composition across agents yields interaction dynamics consistent with the factorized multi-agent updates analyzed in earlier sections.

\subsection{Sequential Text Dynamics and Emergence}

A text-based multi-agent system therefore implements an iteration of the form
\[
\text{dialog}^{(k+1)}
=
\mathcal{G}_m \circ \mathcal{G}_{m-1} \circ \cdots \circ \mathcal{G}_1
\bigl(\text{dialog}^{(k)}\bigr),
\]
where \(\mathcal{G}_i\) denotes the text-generation operator of agent \(i\). Under the encoding \(E\), this corresponds approximately to
\[
x^{(k+1)} \approx T_m \circ T_{m-1} \circ \cdots \circ T_1 \bigl(x^{(k)}\bigr).
\]

Emergent solutions correspond to dialog states that are stable under further agent interaction. These are states for which successive agent updates produce no substantial change. The success of such systems therefore derives not from any single agent optimizing a global objective, but from the structure of factorized interaction that makes latent solution states dynamically accessible.

This provides a principled bridge between idealized operator-theoretic analysis and real multi-agent language systems, without requiring access to internal model representations.

\section{Implications and Limits}

This section discusses implications suggested by the operator-theoretic analysis and clarifies the limits of what is formally established. The focus is on consequences that follow from the structure of the dynamics, rather than on empirical performance claims or implementation-specific heuristics.

\subsection{Implications for Multi-Agent System Design}

The analysis highlights the importance of \emph{constraint factorization} as a structural principle for multi-agent systems. In the formalism studied here, invariant solution structures arise from the interaction of distinct constraint-enforcement operators, rather than from the expressive power or stylistic diversity of individual agents.

From this perspective, agent differentiation is effective insofar as agents are associated with distinct families of constraints whose interaction defines nontrivial invariant sets. Personas can therefore be understood as mechanisms for partitioning evaluative authority, rather than as ends in themselves. When agents enforce identical or nearly identical constraints, the resulting dynamics may reduce to effectively redundant updates, limiting the emergence of new invariant structure.

The theory also clarifies that, under idealized assumptions (e.g., exact projections or proximal updates and sufficient coverage of constraints), variations in update ordering or scheduling affect transient behavior and convergence rates but not the underlying invariant sets. This mirrors classical results for cyclic projection and proximal methods \citep{BauschkeBorwein1996,BauschkeCombettes2017,Boyd2011ADMM}.

\subsection{Limits of the Formalism}

The analysis also makes clear several intrinsic limitations.

First, if the collective feasible set is empty,
\[
\bigcap_{i=1}^m A_i = \varnothing,
\]
then no invariant solution exists, and convergence to a stable solution is impossible. This is a direct consequence of the operator-theoretic framework and holds independently of implementation details.

Second, the results rely on idealized assumptions about constraint enforcement. When updates are noisy, approximate, or inconsistently applied - as in practical systems - the guarantees weaken, and invariant structures may be only approximately realized.

Finally, the formalism characterizes asymptotic behavior and invariant structure, not finite-time performance or robustness. Effects such as dominance of particular agents, imbalance among constraints, or slow convergence due to near-inconsistency are not ruled out by the theory and require separate analysis or empirical investigation.

Accordingly, the framework should be understood as describing what kinds of solution structures are \emph{dynamically accessible in principle}, rather than as a complete account of performance in deployed multi-agent systems.

\section{Conclusion}

This work provides a rigorous explanation for why multi-agent systems composed of large language models can outperform single-agent inference under identical informational conditions. Modeling agents as constraint-enforcement operators acting on a shared state shows that multi-agent interaction implements a factored operator whose invariant sets correspond to intersections of agent-specific constraint families.

We demonstrate that these invariant solution structures are generally not dynamically accessible to any individual agent acting alone, even when all constraints are jointly known. The advantage of multi-agent systems therefore arises not from information aggregation or stochastic diversity, but from operator factorization, which alters the dynamics of inference and stabilizes latent solution structure. Extending the analysis to soft constraint enforcement via proximal operators shows that this effect is robust to approximate and incremental reasoning.

In this sense, emergence does not denote the creation of new solutions, but a change in dynamical accessibility. The solutions reached by multi-agent systems are latent in the shared state space yet inaccessible under unfactored dynamics, becoming reachable only through structured interaction. By grounding multi-agent language model behavior in operator theory and constrained optimization, this work establishes a principled foundation for analyzing and designing multi-agent systems as tractable computational objects rather than heuristic assemblies.

\section*{Acknowledgments}

The author acknowledges the use of large language model-based research assistants for iterative discussion and clarification of mathematical frameworks during the development of this work. All theoretical claims, interpretations, and conclusions are the sole responsibility of the author.

\bibliographystyle{plainnat}
\bibliography{references}

\end{document}